\documentclass[10pt,twocolumn,letterpaper]{article}

\usepackage{iccv}
\usepackage{times}
\usepackage{epsfig}
\usepackage{graphicx}
\usepackage{amsmath}
\usepackage{amssymb}
\usepackage{enumitem}

\newcommand{\argmin}{\mathop{\mathrm{argmin}}}   
\usepackage{tabu}
\usepackage{multirow}
\usepackage{comment}
\usepackage[ruled,vlined]{algorithm2e}
\usepackage[utf8x]{inputenc}
\newcommand*{\our}{GeS\@\xspace}
\newcommand*{\ourOS}{GeOS\@\xspace}

\newcommand*{\auxblock}{\Lambda\@\xspace}
\usepackage{sectsty}

\usepackage{pifont}
\usepackage{gensymb}
\usepackage[export]{adjustbox}
\usepackage{verbatimbox}

\usepackage[pagebackref=true,breaklinks=true,letterpaper=true,colorlinks,bookmarks=false]{hyperref}

\iccvfinalcopy 

\def\iccvPaperID{4832} 

\ificcvfinal\fi
\begin{document}

\title{Learning to Generalize One Sample at a Time with Self-Supervision}

\author{Antonio D'Innocente$^{1,2}$ \hspace{5mm} Silvia Bucci$^{2,3}$ \hspace{5mm} Barbara Caputo$^{2,3}$ \hspace{5mm} Tatiana Tommasi$^{2,3}$\\  
$^1$Sapienza University of Rome, Italy  \hspace{5mm} $^2$Italian Institute of  Technology \hspace{5mm} $^3$Politecnico di Torino, Italy\\ 
\tt\small  \hspace{-7mm} dinnocente@diag.uniroma1.it \hspace{5mm} silvia.bucci@iit.it  \hspace{5mm}\{barbara.caputo,  tatiana.tommasi\}@polito.it }
\maketitle

\begin{abstract}
Although deep networks have significantly increased the performance of visual recognition methods, it is still challenging to achieve the robustness across visual domains that is necessary for real-world applications. To tackle this issue, research on domain adaptation and generalization has flourished over the last decade. An important aspect to consider when assessing the work done in the literature so far is the amount of data annotation necessary for training each approach, both at the source and target level. In this paper we argue that the data annotation overload should be minimal, as it is costly. Hence, we propose to use self-supervised learning to achieve domain generalization and adaptation. We consider learning regularities from non annotated data as an auxiliary task, and cast the problem within an Auxiliary Learning principled framework. Moreover, we suggest to further exploit the ability to learn about visual domains from non annotated images by learning from target data while testing, as data are presented to the algorithm one sample at a time. Results on three different scenarios confirm the value of our approach.  
\end{abstract}

\section{Introduction}
As visual recognition algorithms get ready to be deployed in several markets, the need for 
tools to ensure robustness across various visual domains becomes more pressing. Even when 
massive amounts of data are available, the underlying distributions of training (\ie source) 
and test (\ie target) data are inevitably going to be different. 
Research in the area of adaptive learning has addressed this general issue in various sub-cases, 
from early works on semi-supervised Domain Adaptation (DA) \cite{Saenko:2010,KulisSD11}  
up to very recent attempts to deal with Domain Generalization (DG) \cite{Li_2018_CVPR,hospedales19} 
(for a comprehensive review see section \ref{sec:related}).


\begin{figure}[t]
\centering
\includegraphics[width=0.45\textwidth]{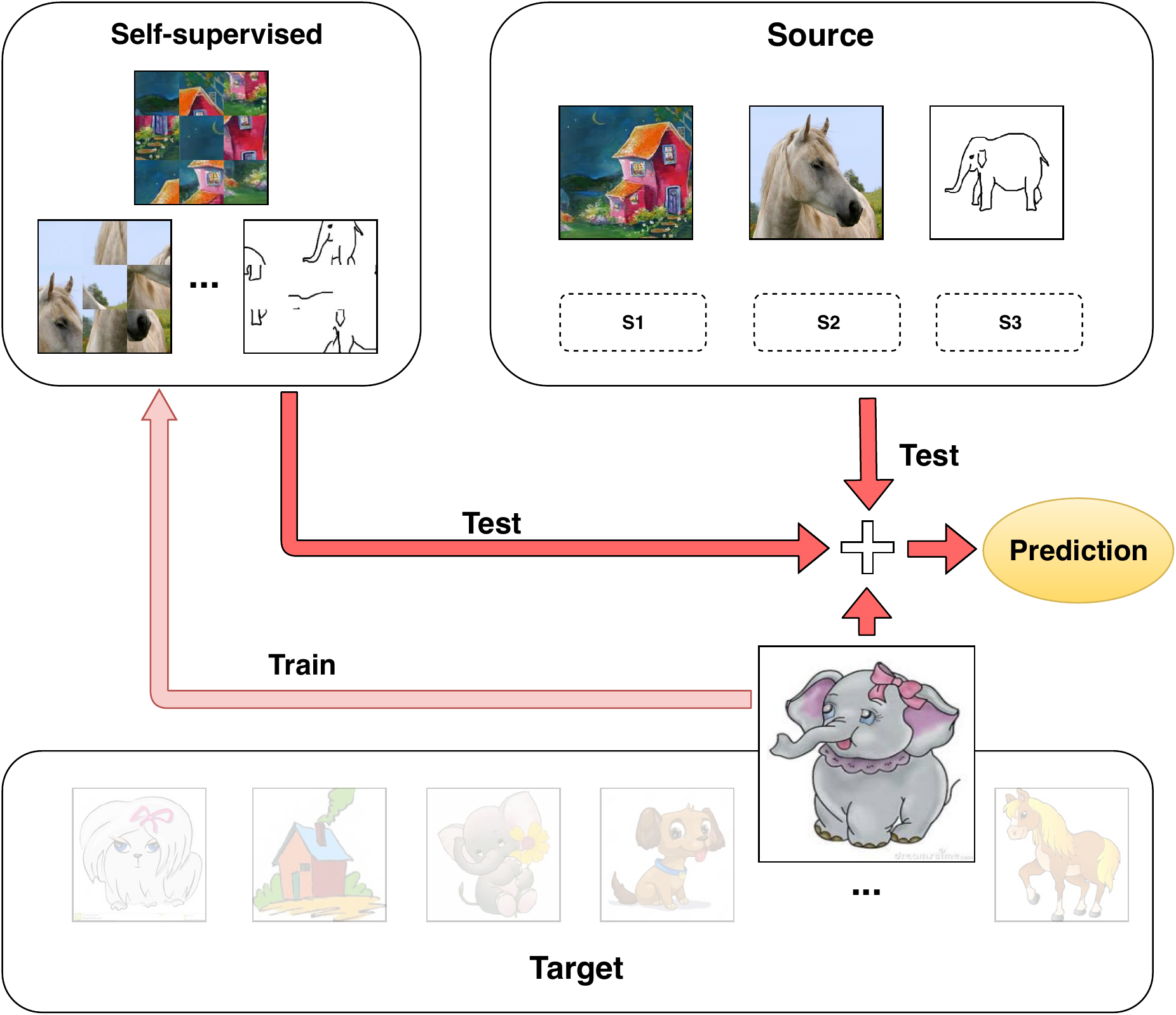}   
\caption{In our approach we propose to exploit self-supervision as an auxiliary task together with the primary supervised task.
Both are learned over multiple source data regardless of their exact domain label (no need to separate S1 from S2 and S3). Incoming unlabeled target samples actively adapt a self-supervised feature extractor module with finetuning. The refined representation is then aggregated with source-based knowledge for the final label prediction.}
    \label{fig:teaser}
\end{figure}

\begin{figure*}[t]
\centering
\includegraphics[width=\textwidth]{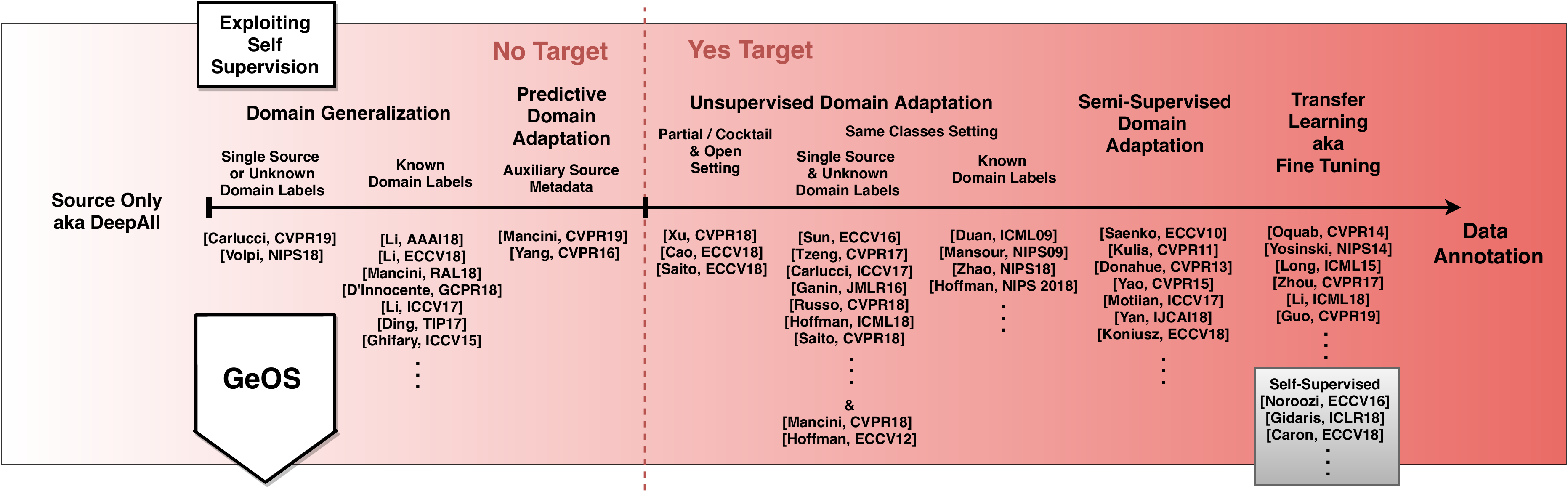}   
\caption{Overview of DA and DG most recent literature sorted on the basis of the
amount of data annotation needed.}
    \label{fig:dataknowledge}
\end{figure*}

An important aspect that remains to be evaluated is the real data annotation effort that is
still needed by the existing DA and DG methods.
Given that the minimum amount of labeling for a multi-class categorization approach 
corresponds to the class identity of the training images (\emph{Source Only} in Figure \ref{fig:dataknowledge}, left), we see that most of the DG algorithms require training data to be annotated also with respect to  
their source domain labels \cite{Li_2018_CVPR,hospedales19}. Approaches proposing to leverage 
over unlabeled auxiliary domains require metadata describing their structure and their relation to the labeled 
source \cite{adagraph}.
DA algorithms need advanced access to large quantity of images from the target domain, 
all depicting the very same classes imaged in the source data, with few notable exceptions
\cite{PADA_eccv18,Saito_2018_ECCV,cocktail_CVPR18}; 
and so forth, up to transfer learning techniques where the source model is useful only
in relation to a good amount of labeled target data, big enough to allow CNN training 
convergence (Figure \ref{fig:dataknowledge}, right).
All these types of data knowledge reveal that the entry point of the current state of the art
algorithms asks for an annotation effort that might still be too costly from the point of view 
of an end user.
Starting from this scenario, the goal of our work is to push visual recognition 
yet one step closer to deployment in the wild.
We aim for a principled method able to generalize to new domains by using only the source class
annotation (no source domain labels) and that, given a single unlabeled target sample at test time, 
 can leverage over its inherent knowledge for a further training before the final prediction
(see Figure \ref{fig:teaser}).
To do that, we exploit self-supervised data knowledge to regularize the learning process of a
source classification model. Similarly to \cite{jigen}, we take into consideration the spatial 
co-location of patches for an image decomposed and reorganized as in a jigsaw puzzle. However,
rather than using a flat multi-task architecture, we design a residual block that focuses
on self-supervised information and provides the main fully-supervised learning flow with
useful complementary knowledge (see Figure \ref{fig:geos}). This strategy has two main advantages.
On one side it improves the stability of the results, removing the need for further control conditions 
on the classification model such as the introduction of an entropy loss in the DA setting as in \cite{jigen}, which requires
a dedicated tuning process for its relative weight. On the other side, by concentrating
the use of self-supervision into a specific part of the network rather than having it distributed, we can easily fine-tune only the auxiliary model at test time on each single 
test sample. 

Summarizing, the contributions of this paper are:
\begin{itemize}[leftmargin=*]
  \item we introduce a new end-to-end deep learning algorithm able to 
  Generalize One Sample at a time (\ourOS) using the same amount of data 
  annotation as the na\"ive \emph{Source Only} baseline.\vspace{-2mm}
  \item we show how self-supervised knowledge can be used in a principled auxiliary
  learning framework for DG and DA, improving in robustness and performance over the
  existing flat multi-task approach \cite{jigen}.\vspace{-2mm}
  \item we present a new generalization setting where it is possible to 
  further train the learning model over every single test sample
  by exploiting self-supervision.\vspace{-2mm}
  \item we show how to get top results in the predictive DA setting \cite{adagraph}
  without the need of human annotated auxiliary knowledge.
\end{itemize}

\section{Related Work}
\label{sec:related}

\paragraph{Self-Supervised Learning}
SSL is a framework developed to learn visual features from large-scale unlabeled data \cite{SSLsurvey}. 
Its first step is the choice of a \emph{pretext} task that exploits inherent data attributes
to automatically generate data labels. It has been shown that the semantic knowledge captured by the first
layers of a network solving those tasks defines a useful initialization for new learning problems.
Indeed the second SSL step consists in transferring the self-supervised learned model of those
initial layers to a real \emph{downstream} task (\eg classification, detection), 
while the ending part of the network is newly trained. 
The advantage provided by the transferred model generally gets more evident, as the 
number of annotated samples of the downstream task is low. 

The pretext tasks can be organized in three main groups. 
One group rely only on original visual cues and involves either the whole image with
geometric transformations (\eg translation, scaling, rotation \cite{gidaris2018unsupervised,NIPS2014_geometric}), 
clustering  \cite{caron2018deep}, inpainting \cite{pathakCVPR16context} and colorization \cite{zhang2016colorful},
or considers image patches focusing on their equivariance (learning to count \cite{learningtocount})
and relative position (solving jigsaw puzzles \cite{NorooziF16,Noroozi_2018_CVPR}).
A second group uses external sensory information either real or synthetic: this solution
is often applied for multi-cue (visual-to-audio \cite{audiovisual}, RGB-to-depth \cite{ren-cvpr2018}) 
and robotic data \cite{grasp2vec, visiontouch}.
Finally, the third group relies on video and on the regularities introduced by the temporal dimension 
\cite{Wang_UnsupICCV2015,SSLvideo}.
The most recent SSL research trends are mainly two. On one side there is the proposal of novel 
pretext tasks, compared on the basis of their ability to initialize a downstream task with 
respect to using supervised models as in standard transfer learning  \cite{OquabTL,YosinskiNIPS2014,FTmedical,SpotTune,Long:2015,liICML18}. 
On the other side there are new approaches to combine multiple pretext tasks together in 
multi-task settings \cite{multitaskSSL,ren-cvpr2018}. 

\paragraph{Domain Adaptation and Generalization}
To cope with domain shift, several algorithms have been developed mainly in two different settings. 
In DA the learning process has access to the labeled source data and to the unlabeled target data, 
thus the aim is to generalize to that specific target set \cite{csurka_book}. The semi-supervised
DA case considers also the availability of a limited number of annotated target samples \cite{Saenko:2010,KulisSD11,doretto2017,semisup1,semisup2,ijcai2018,museumECCV18}.
In DG the target is unknown at training time: the learning process can usually leverage on 
multiple sources to define a model robust to any new, previously unseen domain 
\cite{shallowDG}. In both DA and DG, the main assumption is that source and target share the 
same label set, with very few works studying exceptions to this basic condition 
\cite{PADA_eccv18, cocktail_CVPR18,Saito_2018_ECCV}. 

\emph{Feature-level} strategies focus on learning domain invariant data representations mainly by minimizing 
different domain shift measures \cite{Long:2015,LongZ0J17,dcoral,hdivergence}. The domain shift can also be 
reduced by training a domain classifier and inverting the optimization to guide the features towards maximal domain confusion \cite{Ganin:DANN:JMLR16,Hoffman:Adda:CVPR17}. This adversarial approach has several variants, some of which also exploit class-specific domain recognition modules \cite{saito2017maximum,Li_2018_ECCV}. Metric learning \cite{doretto2017}  and deep autoencoders \cite{DGautoencoders,Li_2018_CVPR,Bousmalis:DSN:NIPS16} have also been used to search for domain-shared embedding spaces. In DG, these approaches leverage on the availability of multiple sources and on the access to the domain label for each sample, meaning that the identity of the source distribution from which every sample is drawn is strictly needed.

\emph{Model-level} strategies either change how the data are loaded with 
ad-hoc episodes \cite{hospedales19}, or modify conventional learning algorithms to search for 
more robust minima of the objective function \cite{MLDG_AAA18}, or introduce domain alignment 
layers in standard learning networks \cite{carlucci2017auto}. 
Those layers can also be used in multi-source DA to evaluate the relation between the sources 
and target and then perform source model weighting \cite{MassiRAL,cocktail_CVPR18}.
Several DG approaches aim at identifying and neglecting domain-specific signatures from multiple 
sources both through shallow and deep methods that exploit multi-task learning \cite{ECCV12_Khosla}, 
low-rank network parameter decomposition \cite{hospedalesPACS,Ding2017DeepDG} or aggregation layers \cite{Antonio_GCPR18,hospedales19}. In multi-source DA 
the domain label of the sources may be unknown \cite{mancini2018boosting, hoffman_eccv12,carlucci2017auto}, while for the DG it 
remains a crucial information that has to be provided since the beginning.

Finally, many recent methods adopt \emph{data-level} solutions based on variants of the Generative Adversarial Networks (GANs, \cite{Goodfellow:GAN:NIPS2014}) to synthesize new images. Indeed, it is possible to reduce the domain gap by producing source-like target images or/and target-like source images \cite{russo17sbadagan,cycada}, as well as a sequence  of  intermediate samples shifting from the source to the target \cite{DLOW}. The data augmentation strategies in \cite{DG_ICLR18,Volpi_2018_NIPS} learn how to properly perturb the source samples, even in the challenging case of DG from a single source. The combination of data- and feature-level strategies has also shown further  improvements in results \cite{ADAGE,sankaranarayanan2017generate}.

Some recent works have started investigating intermediate settings between DA and DG. In \emph{Predictive DA} 
a labeled source and several auxiliary unlabeled domains are available at training time together with metadata 
that describe their relation \cite{adagraph,multivariatereg}.
Other works propose approaches to push model-based DA solutions towards the DG setting adding a memory able 
to accumulate over multiple target samples at test time \cite{adagraph,MassiRAL}. Although it is an interesting 
direction for online and continuous learning, it might only be seen as an upper limit condition to real DG in 
the wild, where we need a separate prediction for every sample.
Moreover, \cite{jigen} has recently started a new research direction moving SSL from the transfer learning
to the domain generalization setting, showing that self-supervision provides useful auxiliary information to 
close the domain gap. In particular it showed that solving jigsaw puzzles improves the generalization 
properties of a supervised classification when both the models are jointly learned with a flat multi-task 
approach.

\paragraph{Multi-Task and Auxiliary Learning}
MTL aims at simultaneously training over several tasks that mutually help each other \cite{CaruanaMTL}. In deep learning this means searching for a single feature representation that works well for multiple tasks.  
This framework is at the basis of many CNN segmentation and detection algorithms \cite{SimultaneousECCV14,FRCNN}.
Several architectures have been investigated to better exploit inter-task connections and task-knowledge complementarity, while growing the number of combined tasks \cite{Kokkinos_2017_CVPR,stitch,AZ_SS}. 
Although powerful, MTL has one main drawback: it is sensitive to the weight assigned to each task, \ie the choice of the
scaling coefficient used to combine multiple loss weights. This causes the need for an extensive 
hyperparameter tuning \cite{Kokkinos_2017_CVPR} 
or for principled loss weighting strategies. Some recent approaches leverages on the evaluation of task uncertainty
\cite{fullyadaptive,kendall2017multi} and dynamically adjust the weights \cite{gradnorm,Guo_2018_ECCV}.

In many real applications the tasks are not all equally important and some prior knowledge on their
ranking is available. In particular, the case with one main \emph{primary} and several other 
\emph{auxiliary} tasks is known as \emph{Auxiliary Learning} (AL) and is 
related to the literature on learning with priviledged information \cite{LUPI}. Very recently \cite{NIPS2018ROCK} presented a residual strategy to integrate multi-modal auxiliary tasks and improve the performance of the primary object detection task. In \cite{auxtasks} the main focus is in the choice of auxiliary tasks which 
should be as cheap as possible in terms of annotation and learning effort.
This research direction is currently attracting more and more attention with also the introduction of unsupervised \cite{aux2} and self-supervised \cite{aux3} auxiliary tasks.


\section{Generalize from One Sample}
\label{sec:approach}
\begin{figure*}[t]
\hspace{1.3cm}\includegraphics[width=0.9\textwidth]{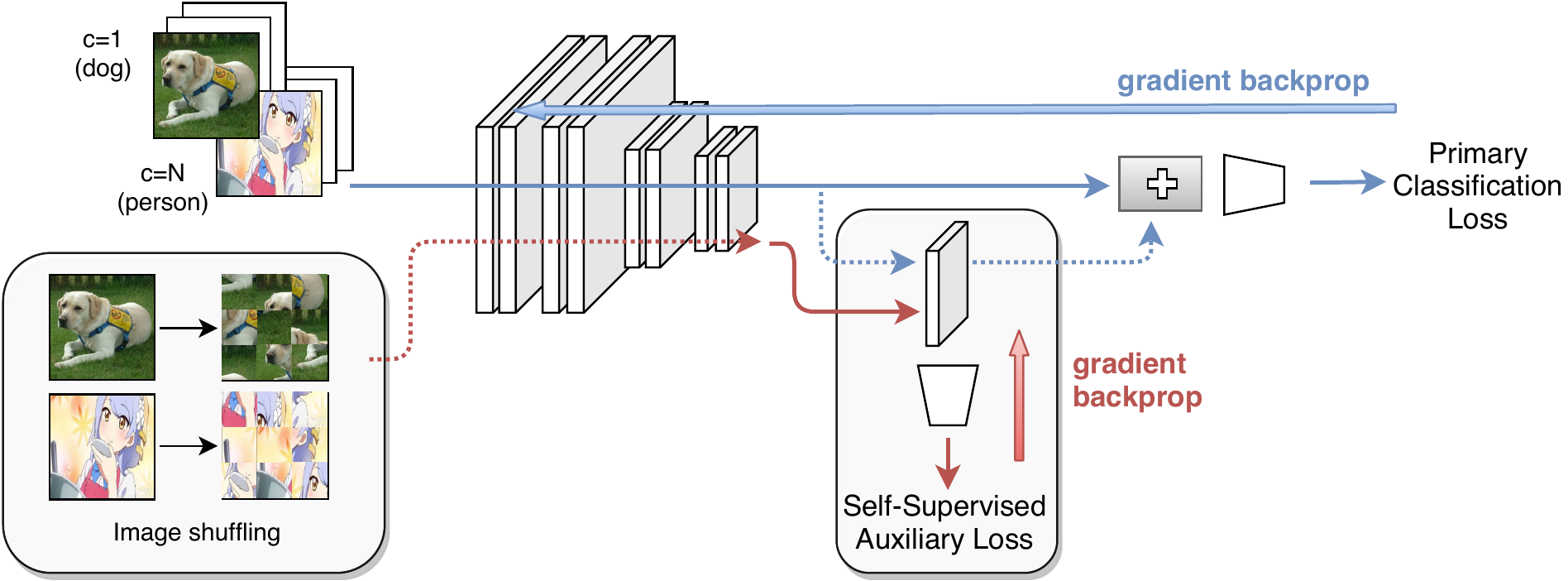}   
\caption{Schematic illustration of \our architecture for learning with self-supervised auxiliary information. The primary network is trained on the supervised task. The auxiliary component refines features for the main classifier, while being trained to solve the Jigsaw Puzzle problem. Lines indicate feature paths in the network. A dotted line means gradients won't be computed for the underlying layers.}
    \label{fig:geos}\vspace{-3mm}
\end{figure*}

The standard DG problem setting considers $i=1, \ldots, S$ source domains, each with $j=1 \ldots N^i$ 
image-label pairs $(x_j^i,y_j^i)$. The goal is learning a model $\mathcal{P}:x\rightarrow y$ that 
generalizes to any test sample drawn from a new target. 
The source domain index $i$ is needed by most of the existing DG algorithms, which use it to 
separate source-specific from source-generic knowledge. We choose instead to ignore it and deal directly 
with samples $(x_j,y_j)$ with $j=1,\ldots,J$ where $J=\sum_i N^i$, focusing only on the class 
annotation $y\in\{ 1\ldots C\}$. 
%
Moreover, by operating simple geometric transformations on 
$x_j$, we can get a variety of new versions $\tilde{x}_j^v$, with $v=1,\ldots,V$.
Examples of transformations may be $90$\degree rotations that lead to $V=4$ possible versions of
each sample \cite{gidaris2018unsupervised}, or $n$-patch based decomposition and 
shuffling as in a jigsaw puzzle, that leads to $V=n!$ variants for each sample \cite{jigen}. 
The obtained self-supervised data-label pairs $(\tilde{x}_k,v_k)$ where $k=1,\ldots,K$ with $K=J \times V$, allow to define an auxiliary classification task $\mathcal{A}:\tilde{x}\rightarrow v$ that can be 
trained jointly with the primary one $\mathcal{P}$, improving its generalization effect 
across multiple sources.

\paragraph{Training Process} The general architecture of our model is shown in Figure \ref{fig:geos}. 
It is composed by a main convolutional backbone that extracts the features $\Theta(x)$ from the original images $x$. It 
 then provides them as input to the fully connected module of the primary task, that is in charge of computing the
classification prediction. To this fairly general network, we add a new residual auxiliary block that
deals with self-supervised data-label pairs. We focus on the jigsaw puzzle task, following
the same approach used in \cite{jigen}. In particular, the original images are decomposed through a
regular $3\times3$ grid in $9$ tiles which are then randomly re-assigned to one of the $9$ grid positions (Figure \ref{fig:geos}, bottom left). Out
of all the possible permutations, we considered a set of $V=30$ cases, using the Hamming distance
based algorithm in \cite{NorooziF16}. Thus, the auxiliary block takes as input the features 
extracted by the fully connected part of the main network from the scrambled images $\Theta(\tilde{x})$. It then
 further process them through few extra convolutional layers, before entering the
final fully connected auxiliary classification module that recognizes the puzzle permutation.
We indicate with $\Lambda(\Theta(x))$ the features encoded by the auxiliary block
(from the original images) that contribute back to the primary
task representation. Indeed, the input to the fully connected module of the primary network is the element-wise sum $\Theta(x) + \Lambda(\Theta(x))$.

We underline that, although the primary and the auxiliary tasks share the initial feature
extraction process and present the described final feature recombination point, they are
actually optimized independently. By indicating with $\mathcal{L}_p(\mathcal{P}(x|\Theta,\Lambda),y)$
the cross-entropy loss of the primary task and with $\mathcal{L}_a(\mathcal{A}(\tilde{x}|\Theta,\Lambda),v)$
the cross-entropy loss of the auxiliary jigsaw task, we overall train the network by optimizing 
the two following objectives:
\begin{align}
    \argmin_{\Theta} \mathcal{L}_p(\mathcal{P}(x|\Theta,\Lambda),y), \\ 
    \argmin_{\Lambda} \mathcal{L}_a(\mathcal{A}(\tilde{x}|\Theta,\Lambda),v).
\end{align}
To summarize it in words, the gradients of the auxiliary loss do not backpropagate into the primary network, 
and the gradients of the primary loss affect the auxiliary module only indirectly through 
the update of the initial convolutional part of the main network. 


\paragraph{Testing Process and One Sample Learning} Given a test sample $x^t$ from an unknown target domain we extract both the
primary $\Theta(x^t)$ and the auxiliary $\Lambda(\Theta(x^t))$ features from it to feed the
classification model, get the prediction and check whether the assigned class is correct or not. 
With respect to this na\"ive testing process, the self-supervised nature of the auxiliary task 
gives us the possibility to further learn from the the single available test sample. Indeed 
we can always decompose the sample in patches to create its shuffled variants and further minimize the
auxiliary puzzle classification loss. In this way the auxiliary block is fine-tuned on the single observed example
and we can expect a benefit from recombining the auxiliary features with those of the primary model. The exact procedure of auxiliary learning from one sample at test time is described in Algorithm \ref{alg:os}.

\begin{algorithm}
\SetAlgoLined
\KwData{source trained model, test sample}
 $\Theta$, $\Lambda$ $\leftarrow$ source trained model \\
 \While{still iterations}{
  ($\tilde{x}^t$, $v^t$) $\leftarrow$ generate random self-supervised mini-batch from test sample variants\\
  minimize the loss $\mathcal{L}_a(\mathcal{A}(\tilde{x}|\Theta,\Lambda),v)$\\
  update $\Lambda^*$ \\
 }
 predict label of test sample using $\Theta$ and $\Lambda^*$
 \caption{One Test Sample Learning}
 \label{alg:os}
\end{algorithm}


\paragraph{Implementation Details}
We instantiate the main network backbone as a ResNet18 architecture and use a standard residual block as our auxiliary self-supervised module. Specifically, the auxiliary block implements a fully connected layer after the last convolution for self-supervised predictions. The main network, parametrized by $\Theta$, is initialized with a pre-trained ImageNet model, while for the auxiliary block parametrized by $\Lambda$, we use random uniform weights. The output of the main network and that of the auxiliary block are aggregated with a plain element-wise sum. 
For each training iteration we feed the network with mini-batches of original $x$ and transformed $\tilde{x}$
images using batch accumulation to synchronously update $\Theta$ and $\Lambda$.
Our architecture has a similar structure to the one recently presented in \cite{NIPS2018ROCK}, but we implemented 
a tailored backpropagation policy to keep separated the primary and the auxiliary 
learning process by zeroing the gradients at both the input and output ends of the auxiliary block.

\section{Experiments}
\label{sec:experiments}
\paragraph{Datasets}
The proposed \ourOS algorithm is mainly designed to work in the DG setting with data from multiple
sources, using only the sample category labels and ignoring the domain annotation. In other words,
\ourOS works with the same amount of data knowledge of the na\"ive \emph{Source Only} reference, also known
as \emph{Deep All} since a basic CNN network can be trained on the overall aggregated source samples.

To test \ourOS in the DG scenario we focused on the PACS  dataset \cite{hospedalesPACS} 
that contains approximately 10.000 images of 7 common categories across 4 different domains 
(Photo, Art painting, Cartoon, Sketch) characterized by large visual shifts. We further investigate the
behaviour of \ourOS in the multi-source DA setting with the same dataset, always considering
one domain as target and the other three as sources.

Finally we evaluate \ourOS in Predictive Domain Adaptation (PDA), a particular DG setting 
that has been recently put under the spotlight by \cite{adagraph}. Here a single labeled data 
is available at training time together with a set of unlabeled auxiliary domains which are provided 
together with extra metadata (image timestamp, camera pose, \etc) useful to derive the reciprocal 
relation among the auxilary sets and the labeled source.
For PDA we follow \cite{adagraph}, testing on  CompCars \cite{Yang_2015_CVPR} and Portraits \cite{Ginosar_2015_ICCV_Workshops}. 
The first one is a large-scale dataset composed of 136,726 vehicle photos taken in the space of 11 years (from 2004 to 2015). As in \cite{adagraph}, we selected a subset of 24,151 images organized in 4 classes (type of vehicle: MPV, SUV, sedan and hatchback) and 30 domains obtained from the combination of 
the year of production (range between 2009 and 2014) and the perspective of the vehicle (5 different view points).
The second dataset is a large collection of pictures taken from American high school year-books. The photos cover a time range between 1905 and 2013 over 26 American states. Also in this case we follow \cite{adagraph} for the experimental protocol: we define a gender classification task performed on 40 domains obtained choosing 8 decades (from 1934) and 5 regions (New England, Mid Atlantic, Mid West, Pacific and Southern).

\paragraph{Domain Generalization}
To align our PACS experiments with the training procedure used in \cite{jigen}, we apply random cropping, random horizontal flipping, photometric distortions and resize crops to 222$\times$222 so that we get equally spaced square tiles on a 3$\times$3 grid for the jigsaw puzzle task. We train the network for 40 epochs using SGD with momentum set at 0.9, an initial learning rate of 0.001, a cumulative batch size of 128 original images and 128 shuffled images and a weight decay of $0.0005$. We divide train inputs in 90\% train and 10\% validation splits, and test on the target with the best performing model on the validation set.
By indicating the auxiliary task loss weight with $\alpha$, we achieve the training convergence for the self-supervised task by assigning $\alpha = 2$, and use that value for all our experiments, including DA and PDA settings, without further optimization. We also leave hyperparameters for the one-sample finetuting steps fixed to their initial training values.

The obtained results are shown in Table \ref{table:resultsDG_PACS}, together with several useful baselines. In particular, JiGen \cite{jigen} was the first method showing that self-supervision tasks can 
support domain generalization, while D-SAM \cite{Antonio_GCPR18} and EPI-FCR \cite{hospedales19}
propose networks with domain specific aggregation layers and domain specific models respectively, with the second one introducing also 
a particular episodic training procedure and getting the current DG state of the art on PACS.
DANN \cite{ganin2014unsupervised} exploits a domain adversarial loss to obtain a source invariant 
feature representation. MLDG \cite{MLDG_AAA18} is a meta-learning based optimization method.
We underline that all these baseline, with the notable exception of JiGen, need source data
provided with both class and domain label. On this basis, the advantage that \ourOS shows
with respect to EPI-FCR is even more significant. 
Since also JiGen leverages over self-supervised knowledge, it might benefit of the One Sample 
Learning procedure at test time as in \ourOS. For a fair comparison we used the code provided
by the authors, implementing and running on it our Algorithm \ref{alg:os}. The row \emph{JiGen + OS}
reports the obtained results, showing a small advantage over the original JiGen, confirming
the beneficial effect of the fine tuning procedure. However the gain is still limited with respect
to the top result of \ourOS: the flat multi-task architecture of JiGen implies a re-adaptation
of the whole network which might be out of reach with a single target sample.
This confirms the effectiveness of the chosen auxiliary learning structure chosen for \ourOS.

\begin{table}[tb]
\begin{center} \small
\begin{tabular}{@{}c@{}c@{}c@{~}c@{~~~}c@{~~~}c|@{~~~}c}
\hline
\multicolumn{2}{c}{\textbf{PACS-DG}}  & \textbf{art\_paint.} & \textbf{cartoon} &  \textbf{sketches} & \textbf{photo} &   \textbf{Avg.}\\ \hline
\multicolumn{7}{c}{\textbf{Resnet-18}}\\
\hline
 \multirow{2}{*}{\cite{Antonio_GCPR18}} & Deep All & 77.87 & 75.89 & 69.27 &  95.19 & 79.55\\
 & D-SAM & 77.33 & 72.43 & 77.83 & 95.30 & 80.72\\
\hline
\multirow{4}{*}{\cite{hospedales19}}& Deep All & 77.60  & 73.90  & 70.30  & 94.40  & 79.10 \\
& DANN & 81.30  & 73.80  & 74.30  & 94.00  & 80.08 \\
& MLDG & 79.50  & 77.30  & 71.50  & 94.30  & 80.70 \\
& EPI-FCR & 82.10  & 77.00  & 73.00  & 93.90  & 81.50 \\
\hline
\multirow{2}{*}{\cite{jigen}} & Deep All & 77.85  & 74.86  & 67.74  & 95.73  & 79.05 \\
 & JiGen    & 79.4  & 75.25  & 71.35  &  96.03 &  80.51\\
 & JiGen \textbf{+ OS}   & 79.40  & 75.24  & 72.26  & 96.27  & 80.79 \\
 \hline
 & \textbf{\ourOS} & 79.79  & 75.06  & 76  & 96.65 & \textbf{81.88} \\
 \hline
\end{tabular}
\caption{Domain Generalization results on PACS. The results of \ourOS are average over 3 repetitions of each run. Each column title indicates the name of the domain used as target.}
\label{table:resultsDG_PACS}
\end{center}
\end{table}

\begin{table}[tb]
    \begin{center} \small
        \begin{tabular}{@{}c@{}c@{}c@{~}c@{~~~}c@{~~~}c|@{~~~}c}
        \hline
        \multicolumn{2}{c}{\textbf{PACS-DG}}  & \textbf{art\_paint.} & \textbf{cartoon} &  \textbf{sketches} & \textbf{photo} &   \textbf{Avg.}\\ \hline
        \multicolumn{7}{c}{\textbf{Resnet-18}}\\
        \hline
        & null hypothesis & 79.26  & 74.09  & 70.13  & 96.23  & 79.93 \\
        & \textbf{\our}    & 78.95   &  74.36 & 73.99 & 96.29 &  80.90\\
        \hline
        & $\textbf{\ourOS}_{it=1}$   &  79.74  & 74.84  & 75.35  & 96.53 & 81.62 \\
        & $\textbf{\ourOS}_{it=2}$   & 79.79  & 75.01  & 76  & 96.61 & 81.85 \\
        & $\textbf{\ourOS}_{it=3}$   & 79.79  & 75.06  & 76  & 96.65 & \textbf{81.88} \\
        \hline
        & $\textbf{\our}_{rotation}$ & 79.49  & 74.11  & 70.6  & 95.87  & 79.52 \\
        & $\textbf{\ourOS}_{rotation}$   & 78.19  & 74.81  & 71.63  & 95.79 & 80.11 \\
        \hline
        \end{tabular}
        \caption{Analysis of several variants of \ourOS: not using the auxiliary knowledge, 
        turning off the one sample finetuning at test time, increasing the number of 
        self-supervised iterations on the target sample and also changing the self-supervised task from
        solving jigsaw puzzles to image rotation recognition.}
        \label{table:resultsDG_PACS2}
    \end{center}
\end{table}

\paragraph{Analysis and Discussion}
We provide a further in-depth analysis of the proposed method, starting from the results in Table  \ref{table:resultsDG_PACS2}. First of all we trained the same network architecture of \ourOS but
without using the auxiliary self-superivised data: in this case we start from the same hyperparameter
initialization setting used for \ourOS but we turn on the gradient propagation over the auxiliary network
block which now behaves as an extra residual layer for the main primary model. The row \emph{null hypothesis}
in the table indicates that the advantage of \ourOS is not due to the increased depth and parameter count,
but originates instead from the proper use of self-supervision and one sample fine tuning. 
To even decouple these last two components, we turn off the one sample learning procedure at test time:
the obtained version \our of our algorithm still outperform JiGen and many of the other competitive
methods in Table \ref{table:resultsDG_PACS}, that yet use more data annotation.

When the fine tuning procedure on the test sample is on, it is possible to optimize the
auxiliary network block with a different number of SGD iterations. We show that the obtained
results increase with the number of iterations, but are already remarkable with a single one.
Finally we evaluate the effectiveness of \ourOS and its simplified version \our when changing
the type of self-supervised knowledge used as auxiliary information. Precisely we follow 
\cite{gidaris2018unsupervised} and rotate the images at steps of 90\degree, training the auxiliary 
block for recognition among the four orientations. 
In this case \our does not provide any advantage with respect to the null hypothesis baseline.
This reveals that the choice of the self-supervised task influences the generalization
capabilities of our approach, but the possibility to still run fine tuning on every single test 
sample maintains a beneficial effect.

\begin{table}[tb]
\begin{center} \small
\begin{tabular}{@{}c@{}c@{}c@{~~}c@{~~}c@{~~}c|@{~~}c}
\hline
\multicolumn{2}{@{}c@{}}{\textbf{PACS-DA}}  & \textbf{art\_paint.} & \textbf{cartoon} &  \textbf{sketches} & \textbf{photo} &   \textbf{Avg.}\\ \hline
\multicolumn{7}{@{}c@{}}{\textbf{Resnet-18}}\\
\hline
\multirow{3}{*}{\cite{mancini2018boosting}}& Deep All & 74.70 & 72.40 & 60.10 & 92.90 & 75.03\\
& Dial & 87.30 & 85.50 & 66.80 & 97.00 &  84.15 \\
& DDiscovery  &  87.70	& 86.90	& 69.60 & 97.00		& 85.30\\
\hline
\multirow{2}{*}{\cite{jigen}}& Deep All & 77.85  & 74.86  & 67.74  & 95.73  & 79.05 \\
& JiGen &  84.88 &	81.07 &	79.05 &	97.96 &	\textbf{85.74} \\
\hline
& \textbf{\our}    & 80.96  &  77.56 & 78.78  & 97.39 &  83.67\\
\hline
\end{tabular}
\caption{Multi-source Domain Adaptation results on PACS obtained as average over 3 repetitions for each run.}
\label{table:resultsDA_PACS}
\end{center}
\end{table}

\paragraph{Unsupervised Domain Adaptation}
Although designed for DG, our learning approach can also be used in the DA setting.
To test its performance, we run experiments on PACS as already done in \cite{jigen}. 
We choose the same training hyperparameters used in DG experiments, with the difference that we train the self-supervised task using images from the target unlabeled domain only, and we validate the network 
on the self-supervised jigsaw puzzle task using an held-out split from the target. 
Since all the target data are now available at once, the one sample finetuning strategy is superfluous, thus we fall back to the simplified \our version of our approach.
Even just exploiting the self-supervised knowledge and not using any explicit domain adaptation strategy, results in Table \ref{table:resultsDA_PACS} show that \our reduces the domain gap with the target domain, yielding an accuracy increase of more than 4 percentage points over the Deep All baseline. 

\begin{figure}[tb]
\centering
\includegraphics[height=4cm]{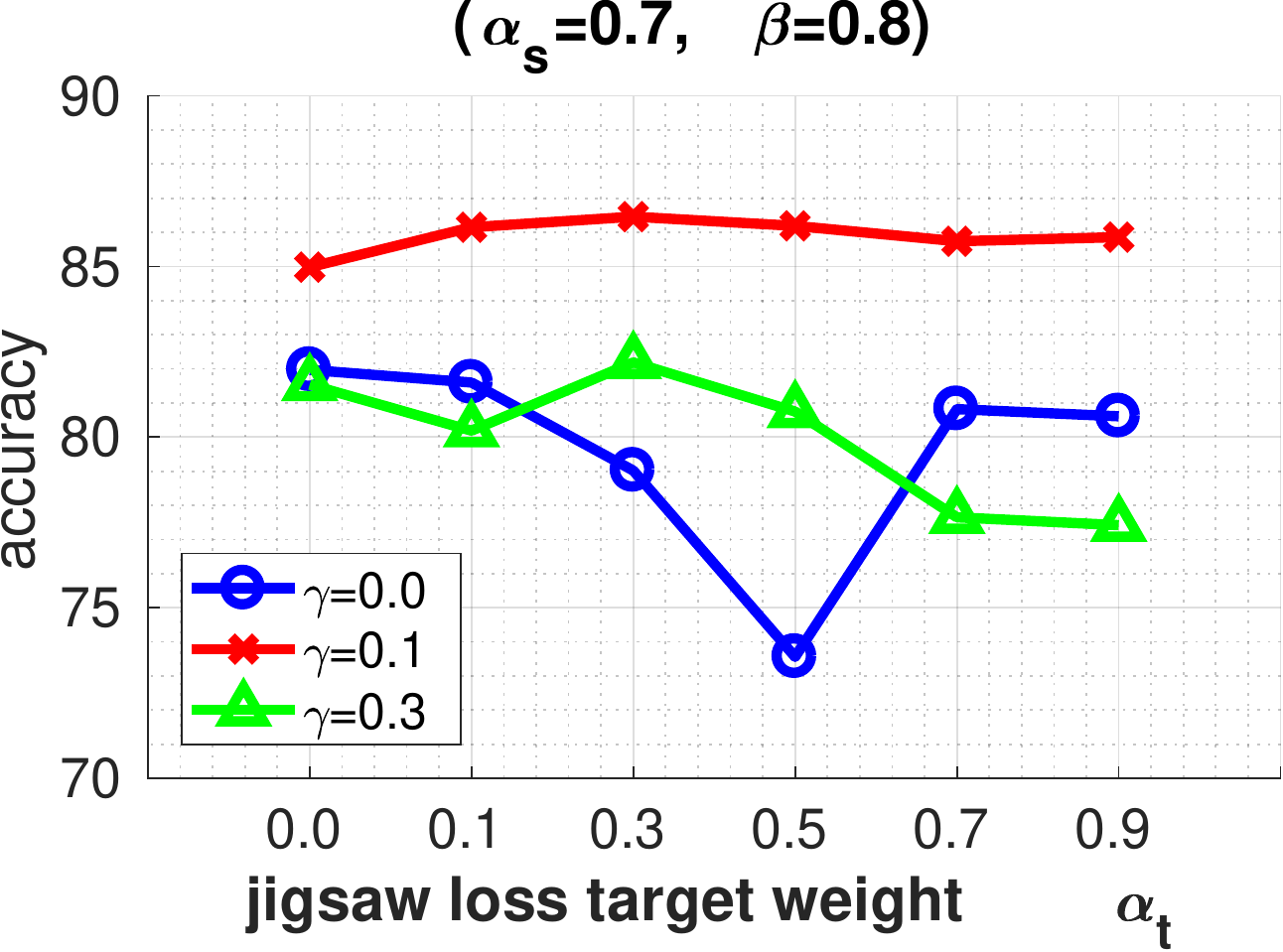}
\caption{Analysis of JiGen in the PACS DA setting. The parameter $\gamma$ weights the 
entropy loss that involves the target data. Moreover the method exploits two different
($\alpha_s, \alpha_t$) auxiliary loss weights related to the self-supervised
task, besides the parameter $\beta$ used to regulate the data loading procedure for 
original and shuffled images.}
    \label{fig:ablation_jigsaw}
\end{figure}

Both DDiscovery \cite{mancini2018boosting} and Dial \cite{carlucci2017auto} are methods that
can be applied on the whole set of source samples without their domain label, as well as JiGen \cite{jigen}, thus
the comparison with \our here is fair in terms of data annotation involved. However, it is useful to remark 
that all those methods minimize an extra entropy loss on the target data. Although
it might be beneficial for adaptation, this further learning condition is not applicable in the DG setting 
and introduces a further computational burden due to the need of tuning the relative loss weight to 
adjust its relevance with respect to the other losses already included in the training model.
For a better understanding, we focus on JiGen and analyze its behaviour when changing the entropy loss
weight $\gamma$. The obtained performance is presented in Figure \ref{fig:ablation_jigsaw} and
clearly indicate that JiGen is fairly sensitive to $\gamma$, besides having overall more ad-hoc
hyperparameters that \our and \ourOS.

\begin{table}[tb]
\begin{center} \small
\begin{tabular}{cccc}
\hline
\multicolumn{4}{@{}c@{}}{\textbf{Resnet-18}}\\
\hline
Method & CompCars & Portraits-Dec. & Portraits-Reg.\\ \hline
Baseline & 56.8 & 82.3 & 89.2\\
AdaGraph & \textbf{65.1} & 87.0 & 91.0\\ \hline
\textbf{\our} & 60.2 & \textbf{87.1} & \textbf{91.6}\\
\textbf{\ourOS} & 60.0 & \textbf{87.1} & 91.5 \\
\hline
\end{tabular}
\caption{Predictive DA results.}
\label{table:resultsCOMPCARS}
\end{center}
\end{table}

\paragraph{Predictive DA without metadata}
The minimal need of supervision of \ourOS puts it in a particularly profitable condition with respect
to other existing DG methods in the challenging Predictive DA experimental setting. Indeed \ourOS can ignore
the availability of metadata and exploit directly the large scale unlabeled auxiliary sources.
We compare the performance of our method against AdaGraph \cite{adagraph}, a very recent approach that
exploits domain-specific batch-normalization layers to learn models for each source domain in a graph,
where the graph is provided on the basis of the source auxiliary metadata.

We follow the experimental protocol described in \cite{adagraph}. For CompCars, we select a pair of domains as source and target and use the remaining 28 as auxiliary unlabeled data. Considering all possible domain pairs, we get 870 experiments and observe the average accuracy results over all of them. 
A similar setting is applied for Portraits, for which we consider the \textit{across decades} scenario (source and target domains selected from the same decade) and the \textit{across region} scenario (source and target from the same region). In total we run 440 experiments for Portraits.

More in details, for CompCars, we start from an ImageNet pretrained model and trained for 6 epochs on source domain using Adam as optimizer with weight decay of $10^{6}$. The batch size used is 16 and the learning rate is $10^{-3}$ for the classifier and $10^{-4}$ for the rest of the network; the learning rate is decayed by a factor of 10 after 4 epochs. 
In the case of Portraits the main learning procedure remains the same used above, except for the number of epochs that in this case is 1 and for the jigsaw weight that in this case was set to $2.0$ for the experiments \textit{across decades} and to $1.0$ for the experiments \textit{across regions}.\\

Table \ref{table:resultsCOMPCARS} show the obtained results, indicating that \our outperforms AdaGraph in all settings except CompCars, despite using much less annotated information. In this particular setting, turning on the fine tuning
process on a each target sample is irrelevant: the amount of auxiliary source data is so abundant that 
the self-supervised auxiliary task is already providing its best generalization effect, thus \ourOS does not show any further advantage with respect to \our.

\section{Conclusions}
This paper presented the first algorithm for domain generalization  able to learn from target data at test time, as images are presented for classification. We do so by learning regularities about target data as auxiliary task through self-supervision. The algorithm is very general and can be used with success in several settings, from classic domain adaptation to domain generalization, up to scenarios considering the possibility to access side domains \cite{adagraph}. Moreover, the principled AL framework leads to a notable stability of the method with respect to the choice of its hyperparameters, a highly desirable feature from deployment in realistic settings. Future work will further investigate this new generalization scenario, studying the behaviour of the approach with respect to the amount and the quality of  unsupervised data available at training time.

{\small
\bibliographystyle{ieee}
\bibliography{egbib}
}

\newpage
\allsectionsfont{\centering}
\section*{Supplementary Material}




We provide here an extended discussion and further evidences of the advantage introduced by learning to generalize
one sample at a time through the proposed auxiliary self-supervised finetuning process.
First of all we clarify the difference between our full method named \ourOS and its
simplified version \our. 

\vspace{2mm}
\textbf{\our} is the architecture we designed for deep learning \textbf{Ge}neralization by exploiting 
\textbf{S}elf-Supervision. Its structure is depicted in Figure 3 of the paper.
Besides the main network backbone that tackles the primary classification task, we introduce an
auxiliary block that deals with the self-supervised objective. It provides useful complementary features 
that are finally recombined with those of the main network improving the robustness of the primary model. 

We mostly focused on the jigsaw puzzle self-supervised task, thus our auxiliary data are scrambled version of 
the original images, recomposed with their own patches in disordered positions. 
This specific formalization for jigsaw puzzle was recently introduced in \cite{jigen}, where
the method JiGen learns jointly over the ordered and the shuffled images with a flat multi-task 
architecture.
Although it showed to be effective, this approach substantially disregards the warnings highlighted 
in \cite{NorooziF16} about the need of avoiding shortcuts that exploit low-level image statistics 
rather than high-level semantic knowledge while solving a self-supervised task. 
By fitting the self-supervised knowledge extraction in an auxiliary learning framework, \our keeps the beneficial 
effect provided by self-supervision without the risk of confusing the learning process with  
low-level jigsaw puzzle specific information. Indeed the auxiliary knowledge is extracted by a 
dedicated residual block towards the end of the network together with a tailored backpropagation strategy 
that keeps the primary and the auxiliary tasks synchronized but separated in their specific
objectives.

In the DA setting, the auxiliary block of \our is trained exclusively by the target images, that 
are available at training time but are unlabeled. In this case when the ordered source images enter the
auxiliary block we obtain target-style-based features that allow to bridge the gap across
domains. Indeed, these features are recombined with the ones from the main backbone and together guide 
the learning process of the primary classification model.
In DG, the ordered source images are provided
as input to the primary classification task, while their scrambled versions are fed to the auxiliary block. Thus the scrambled source images train the auxiliary block, that is finally used as a complementary feature extractor for the respective ordered images. 

At test time, for DA each of the ordered target images pass both through the main and through the 
auxiliary block for feature extraction using the network model obtained at the end of the training
phase.
In DG, for each target sample we may follow the same procedure used for DA testing. 
However, we can also do more by leveraging on self-supervision to distill further knowledge.

\begingroup
\begin{table}[t]
\begin{center} \small
\begin{tabular}{@{~}c@{~~}c@{~~~}c@{~~~~}c@{~~~~}c@{~~~~}c@{~}}
\hline
target                       & run     & \our      & \ourOS$_{it=1}$     & \ourOS$_{it=2}$    & \ourOS$_{it=3}$     \\
\hline
\multirow{3}{*}{photo}       & 1            & 96.41     & +0.12     & +0.18     & +0.30     \\
                             & 2            & 96.41     & +0.30     & +0.48     & +0.36     \\
                             & 3            & 96.05     & +0.30     & +0.30     & +0.42     \\
\hline
\multirow{3}{*}{art\_paint.}      & 1            & 79.00     & +0.74     & +1.13     & +1.37     \\
                             & 2            & 78.71     & +0.83     & +0.33     & +0.09     \\
                             & 3            & 79.15     & +0.78     & +0.64     & +0.64     \\
\hline
\multirow{3}{*}{cartoon}     & 1            & 73.72     & +0.85     & +0.67     & +0.79     \\
                             & 2            & 74.23     & +0.17     & +0.34     & +0.39     \\
                             & 3            & 75.13     & +0,42     & +0.55     & +0.51     \\
\hline
\multirow{3}{*}{sketches}      & 1            & 73.45     & +1.53     & +2.04     & +2.04     \\
                             & 2            & 74.14     & +1.35     & +2.16     & +1.81     \\
                             & 3            & 74.37     & +1.20     & +1.83     & +2.19    \\
\hline
\end{tabular}
\caption{PACS-DG accuracy gains of \ourOS over \our when finetuning separately over each target sample 
at test time with an increasing number of iterations.} \vspace{-8mm}
\label{table:iterations}
\end{center}
\end{table}
\endgroup

\vspace{2mm}
\textbf{\ourOS} is our full method that exploits the  architecture of \our and  runs a fine-tuning procedure at test time for each target sample in the DG setting. The target image is scrambled and provided as input to the auxiliary block which is initialised with the model obtained from the scrambled source images at training time. Although starting from a single instance, the standard data augmentation together with the scrambling procedure provide us with enough samples to fine-tune the auxiliary block. Of course minimizing the jigsaw loss means running multiple SGD iterations for the network parameter updates. 

\vspace{2mm} 
\textbf{Table \ref{table:iterations}}  extends Table 2 of the paper, showing how subsequent iterations of the auxiliary block optimization process always introduces an improvement with respect to \our. We executed three runs for each experiment and the results indicate that the advantage is always present in each single experiment
and it is not just an effect visible on average.
We underline that, although also the method JiGen \cite{jigen} exploits
self-supervised knowledge for domain generalization, it does not consider the possibility 
to adapt the network at test time on each target sample. Indeed, its flat
multi-task structure would imply an overall update of the network, while with \ourOS
we can focus on adapting exclusively the auxiliary knowledge block with a 
larger benefit on the obtained DG accuracy as shown in Table 1 of the 
main submission. 

\end{document}